\pdfoutput=1

\documentclass[11pt]{article}

\usepackage{ACL2024}

\usepackage{latexsym}

\usepackage{multirow}
\usepackage{graphicx}
\usepackage{array}
\usepackage{tabularx}
\usepackage{lipsum}
\usepackage{arydshln}
\usepackage{paralist}

\usepackage{times}
\usepackage{latexsym}

\usepackage[T1]{fontenc}

\usepackage[utf8]{inputenc}

\usepackage{microtype}

\usepackage{inconsolata}
\newcommand{\quotes}[1]{``#1''}

\title{Retrieval augmented text-to-SQL generation for epidemiological question answering using electronic health records}

\author{Angelo Ziletti* \\
  Bayer AG\\
  \texttt{angelo.ziletti@bayer.com} 
\\\And
 Leonardo D'Ambrosi \\
  Bayer AG\\}

\begin{document}
\maketitle
\begin{abstract}
Electronic health records (EHR) and claims data are rich sources of real-world data that reflect patient health status and healthcare utilization. 
Querying these databases to answer epidemiological questions is challenging due to the intricacy of medical terminology and the need for complex SQL queries. Here, we introduce an end-to-end methodology that combines text-to-SQL generation with retrieval augmented generation (RAG) to answer epidemiological questions using EHR and claims data. We show that our approach, which integrates a medical coding step into the text-to-SQL process, significantly improves the performance over simple prompting. Our findings indicate that although current language models are not yet sufficiently accurate for unsupervised use, RAG offers a promising direction for improving their capabilities, as shown in a realistic industry setting. 
\end{abstract}

\section{Introduction} \label{sec:Intro}
Real-world data (RWD) are data routinely gathered from various sources that capture aspects of patient health status and the provision of health care. This encompasses electronic health records (EHR), medical claims data, disease registries, and emerging sources like digital health technologies.
By investigating epidemiological quantities like patients' counts and demographics, disease incidence and prevalence, natural history of diseases, and treatment patterns in real-world clinical practice, researchers and healthcare organizations can identify for example target patient populations with unmet needs, discover unknown benefits of available drugs, evaluate potential for market entry, and estimate the potential enrolment of clinical trials.

\noindent 
\textbf{Problem Statement.}
Addressing epidemiological questions using RWD databases is complex, as it requires not only an understanding of the data's characteristics, including biases, confounders, and limitations, but also involves interpreting medical terminology across various ontologies, formulating precise SQL queries, executing these queries, and accurately synthesizing the results. 

\noindent 
\textbf{Contributions.} 
With this paper, we present a straightforward and effective end-to-end approach to answer epidemiological questions based on data queried from EHR/Claims databases.
\begin{itemize}
\item We release a dataset of manually annotated question-SQL pairs designed for epidemiological research, and adhering to the widely-adopted Observational Medical Outcomes Partnership Common Data Model (OMOP-CDM)~\cite{omop-cdm-2023}.
\item We integrate a medical coding step into the text-to-SQL process, enhancing data retrieval and clinical context comprehension. 
\item We show that retrieval augmented generation (RAG) significantly improves performance compared with static instruction prompting, as confirmed by extensive benchmarking with top-tier large language models (LLMs).
\item We share our dataset, code, and prompts\footnote{\url{https://github.com/Bayer-Group/text-to-sql-epi-ehr-naacl2024}} to foster reproducibility and catalyse a community-driven effort towards advancing this research area.
 \end{itemize}

The presented approach is currently deployed at Bayer in experimental mode. Epidemiologists and data analysts are using the system to explore and evaluate its capabilities, ensuring that its use is carefully monitored and supervised.

\section{The Dataset} \label{sec:dataset}
\begin{table}[htb]
\small
\centering
\begin{tabular}{p{0.25\textwidth}p{0.15\textwidth}}
\hline\hline
\textbf{Quantity} & \textbf{Value}\\
\hline
\# of question/SQL pairs (all)& 306 \\
\# of different tables used (all)& 13 \\
\# of different columns used (all) & 44 \\
\# logical conditions/query & 6.4 (6.7) \\
\# nesting levels/query & 1.5 (1.1) \\
\# tables/query & 2.7 (0.9) \\
\# columns/query & 6.3 (4.7) \\
\# medical entities/query& 2.0 (4.1) \\
Question length [char]/query & 91.7 (81.2) \\
SQL query length [char]/query & 796.4 (448.5) \\
\hline
\hline
\end{tabular}
\caption{Summary statistics of the dataset. For sample statistics, average and standard deviation (in brackets) are reported.}
\label{table:dataset-statistics}
\end{table}

Our dataset was created through a manual curation process, engaging specialists in epidemiological studies to contribute typical questions from their work.
Despite its modest size, the dataset offers a realistic selection of epidemiological questions within industry practice, and exhibits a high degree of complexity. 53 samples require more than two level of nesting, and 19 more than three levels. 
Correctly answering questions often require multiple logical steps: selection of population(s) of interest, relationship between events within a specific time frame, aggregation statistics, and basic mathematical operations (e.g., ratios).
The dataset features questions in their natural, free-form language and it is augmented with two paraphrased versions per question-SQL pair, increasing volume while also offering validated labels for retrieval algorithms. Statistics on the dataset are shown in Table~\ref{table:dataset-statistics}. Due to budget limits, we will use one version per question for subsequent evaluations.

\noindent 
\textbf{Applicability across RWD databases}.
To address the challenge of data retrieval variability across databases with differing data models, we leverage the OMOP-CDM. This model, underpinned by standardized vocabularies~\cite{reich-2024-ohdsi}, harmonizes observational healthcare data and it is widely recognized as the standard for RWD analysis, with data from over 2.1 billion patient records across 34 countries~\cite{voss-etal-2023-european,reich-2024-ohdsi}. 

\begin{figure*}[htb]
\centering
\includegraphics[width=\textwidth]{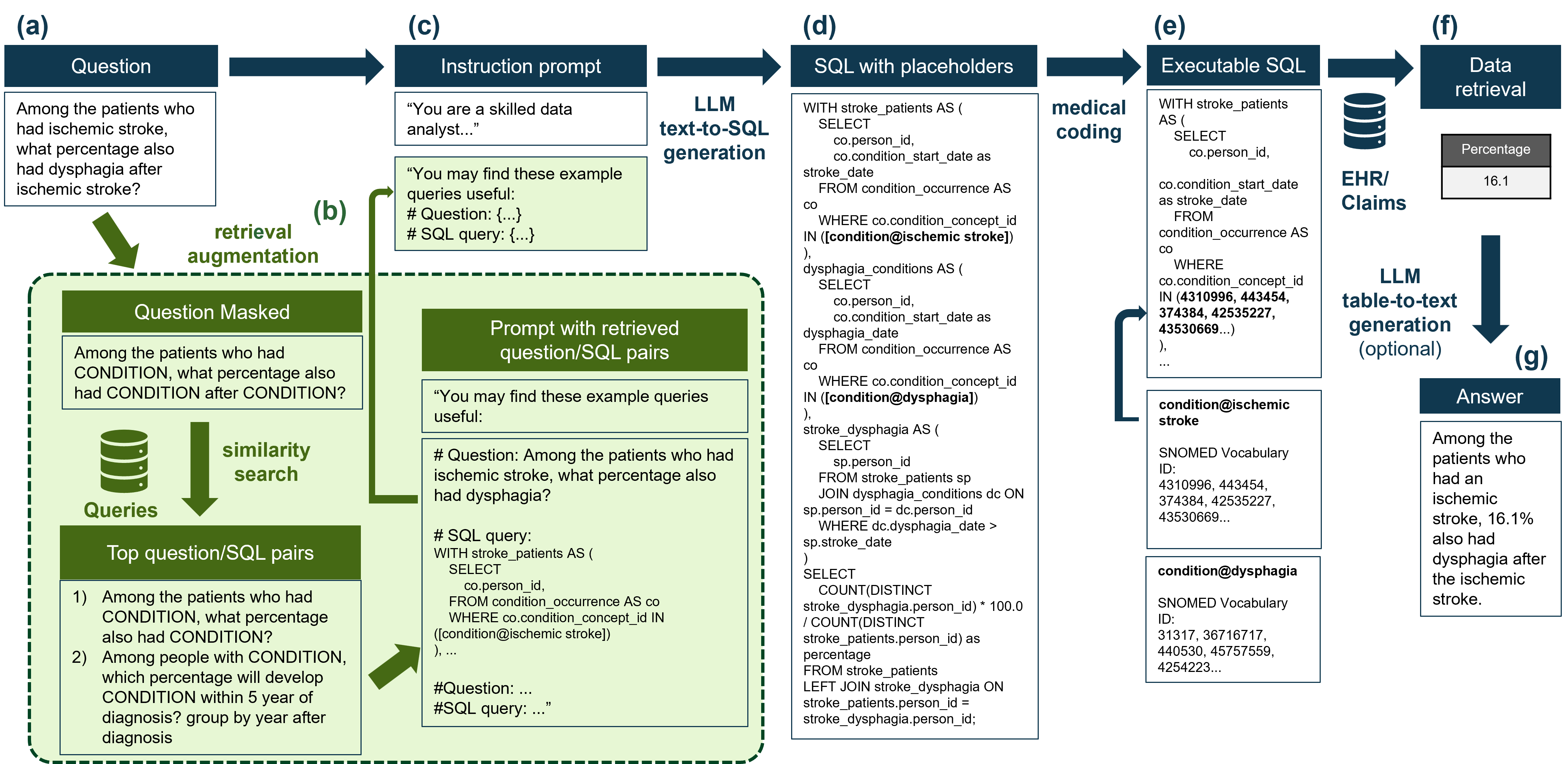}
\caption{From a question in natural language to an answer in natural language using electronic health record or claims databases: end-to-end workflow.}
\label{fig:figure-workflow}
\end{figure*}

\section{Methods} \label{sec:methodology}
Our methodology, outlined in Fig.~\ref{fig:figure-workflow}, employs LLM prompting to translate natural language questions into SQL queries. It advances EHR text-to-SQL methods beyond the constraints of exact or string-based matching to fully encompass the semantic complexities of clinical terminology~\cite{wang-etal-2020-text,lee-2022-ehrsql}.
To achieve this, we introduce a step where an LLM generates SQL with placeholders for medical entities (e.g., [condition@disphagia] in Fig.~\ref{fig:figure-workflow}d), which are then mapped to precise clinical ontology terms (Sec.~\ref{sec:med-coding}, Fig.~\ref{fig:figure-workflow}d-e). This yields executable queries that accurately retrieve database information.
Building on the success of RAG in enhancing LLMs for complex NLP tasks~\cite{lewis-2021-retrieval}, we use our dataset (Sec.~\ref{sec:dataset}) as an external knowledge base. Relevant question-SQL pairs are extracted and incorporated into the prompt, refining SQL generation. The completed SQL queries, embedded with medical codes, are run on an OMOP CDM-compliant database (Fig.~\ref{fig:figure-workflow}f) to facilitate data retrieval. If needed, an answer can be articulated from the retrieved data through further LLM prompting (Fig.~\ref{fig:figure-workflow}g).

\section{Evaluation} \label{sec:evaluation}
\subsection{Experimental setup} \label{sec:exp-setup}

\noindent
\textbf{Large language models.}
We employ several leading LLMs as of February 2024: OpenAI's GPT-3.5 Turbo~\cite{brown-2020-language} and GPT-4 Turbo~\cite{openai2023gpt4}, Google's GeminiPro 1.0~\cite{geminiteam2023gemini}, Anthropic's Claude 2.1~\cite{anthropic2023claude}, and Mistral AI's Mixtral 8x7B and Mixtral Medium~\cite{mistral-medium-2024}, with Mixtral 8x7B being the only open-source model~\cite{jiang2024mixtral}.
We use one simple and one advanced prompt. The simple prompt provides essential instructions for creating queries that adhere to the conventions of the pipeline (Fig.~\ref{fig:figure-workflow}). The advanced prompt adds detailed directives on concept IDs, race analysis, geographical analysis, date filters, column naming, patient count, age calculation, and additional instructions on SQL query validity review.
Following ~\citet{pourreza-2023-dinsql}, we allow LLMs up to three attempts to self-correct non-executable SQL queries using the compiler's error feedback.

\noindent
\textbf{Retrieval augmented generation.}
For similarity computation in RAG, we apply entity masking to substitute medical entities with generic labels (e.g., <DRUG>).
We utilize the \textsc{bge-large-en-v1.5} embedding model from Hugging Face~\cite{wolf-etal-2020-transformers}, which has been fine-tuned for retrieval augmentation of LLMs ~\cite{zhang-etal-2023-retrieve}.
We opt for masked question selection rather than utilizing the query because it eliminates the need for an initial LLM call to generate SQL for retrieval, while maintaining a comparable accuracy~\cite{gao-2023-text}. 

\noindent
\textbf{Medical coding.} \label{sec:med-coding}
LLMs extract medical entities and integrate them into SQL as placeholders (Fig.~\ref{fig:figure-workflow}d), effectively recasting the medical coding task into medical entity normalization ~\cite{portelli-etal-2022-generalizing, ziletti-etal-2022-medical,zhang-etal-2022-knowledge,limsopatham-collier-2016-normalising}. To perform entity normalization, we first compute the cosine similarity of each entity's SapBERT embeddings ~\cite{liu-etal-2021-self} with SNOMED ontology terms, and select the top-50 matches. Then, similarly to ~\citet{yang2022-multilabel}, we prompt GPT-4 Turbo to verify whether a given code should be assigned to the input entity, refining the list.

\noindent
\textbf{Database and evaluation.} \label{sec:database-synpuf}
The evaluation data reported are obtained querying the DE-SynPUF dataset~\cite{synpuf-cms-2008}, which is a synthetic dataset that emulates the structure of actual claims data. It includes 6.8 million beneficiary records, 112 million claims records, and 111 million prescription drug events records \cite{gonzales-2023-synthetic}. The same analysis could be applied to any database conforming to the OMOP-CDM, thus potentially allowing access to 2.1 billion patient records~\cite{reich-2024-ohdsi}.
For evaluation, we manually developed a dataset of question-SQL pairs, as detailed in Sec.~\ref{sec:dataset}. These are then executed against the DE-SynPUF dataset, and the retrieved data from both reference and generated queries are compared to assess performance. This process reflects the practical use of SQL queries on healthcare databases.
To ensure a realistic evaluation setup, the actual question being evaluated is removed from the RAG procedure.
A generated query is marked as correct if it retrieves data enabling an answer that aligns with the reference query's answer (within a 10\% tolerance), and incorrect otherwise. The tolerance compensates for variations from GPT-4 Turbo-based medical coding, maintaining the focus on text-to-SQL evaluation accuracy.

\subsection{Experimental results} \label{sec:exp-results}
Results are shown in Table~\ref{table:evaluation}, and outlined below.
\begin{table*}[htb]
\small
\centering
\begin{tabular}{rllllllllllll}
\hline
\hline
\multicolumn{1}{l}{}                 & \multicolumn{2}{c}{\textbf{Mixtral}}            & \multicolumn{2}{c}{\textbf{GeminiPro}}          & \multicolumn{2}{c}{\textbf{Claude 2.1}}            & \multicolumn{2}{c}{\textbf{Mistral-m}}        & \multicolumn{2}{c}{\textbf{GPT-3.5 Turbo}}              & \multicolumn{2}{c}{\textbf{GPT-4 Turbo}}              \\ \hline
\multicolumn{1}{l}{}                 & \multicolumn{1}{c}{Acc} & \multicolumn{1}{c}{Exec} & \multicolumn{1}{c}{Acc} & \multicolumn{1}{c}{Exec} & \multicolumn{1}{c}{Acc} & \multicolumn{1}{c}{Exec} & \multicolumn{1}{c}{Acc} & \multicolumn{1}{c}{Exec} & \multicolumn{1}{c}{Acc} & \multicolumn{1}{c}{Exec} & \multicolumn{1}{c}{Acc} & \multicolumn{1}{c}{Exec}  \\ \hline
\multicolumn{1}{r}{Prompt (simple)} & 2.0   & 7.8   &  6.9  & 29.4    &   \underline{20.6}     &  53.5     &   8.8    &  32.4  & 20.2   &   \underline{67.0}    & \textbf{28.4}    & \textbf{77.5}   \\
\hdashline[0.5pt/2pt]
\multicolumn{1}{r}{Prompt (advanced)} & 2.9   & 18.6   &  6.9  &  34.7   &   \underline{25.5}     &   \underline{78.4}     &  17.6     & 44.1   &  15.8  &  63.4    & \textbf{38.2}    & \textbf{91.2}   \\
 \hline
\multicolumn{1}{r}{RAG-random1}  & 19.6   & 46.1   &  11.8  &  35.3   &  33.3     &  76.5     &   \underline{38.2}     & 68.3   & 29.0   &   \underline{84.0}    &   \textbf{50.0}  &  \textbf{97.1}  \\
\hdashline[0.5pt/2pt]
\multicolumn{1}{r}{RAG-top1}  & 33.3   &  52.0  &  38.2  &  59.8   &  29.4   &  73.5     &  50.0     &  69.6  &   \underline{59.8}  &  \underline{90.2}     & \textbf{72.5}    &  \textbf{97.1}  \\
\hdashline[0.5pt/2pt]
\multicolumn{1}{r}{RAG-top2} & 20.6   &  40.2  &  37.3  &  56.9   &  38.6     &  75.2     &  46.1    & 73.5   &  \underline{61.8}   &  \underline{94.1   }  &  \textbf{77.5}   & \textbf{98.0}   \\
\hdashline[0.5pt/2pt]
\multicolumn{1}{r}{RAG-top5} &  22.5  &  44.1  &  35.0  &  62.0   &   34.3    & 71.6      &    51.0   &   73.5 &   \underline{52.0}  &  \underline{95.1}    & \textbf{77.5}    &  \textbf{97.1}  \\
\hline
\multicolumn{1}{r}{RAG-top1-oracle} &  52.0  &  62.7  &  67.6  &  73.5   &  58.8     &  83.3     &  56.9     &  74.5  &  \textbf{91.1}  &  \textbf{99.0}    & \underline{82.8}    & \underline{95.0}   \\
\hline \hline
\end{tabular}
\caption{Comparative evaluation of LLMs' performance on text-to-SQL generation for epidemiological question answering. Accuracy (Acc) and executability (Exec) percentages are presented across different models and prompting conditions. Best results are in bold, while second best are underlined. RAG-top1/2/5 indicates the use of the top 1, 2, or 5 most similar questions to augment generation. RAG-random1 and RAG-top1-oracle scenarios provide models with a random dataset sample and the correct SQL query, respectively, for context.}
\label{table:evaluation}
\end{table*}

\noindent 
\textbf{Enhanced performance with detailed prompting.}\label{sec:prompt-simple-adv} 
Advanced prompting typically increases execution scores across models (except GPT-3.5 Turbo), but its impact on accuracy varies: Claude 2.1, Mistral-m, and GPT-4 Turbo show marked accuracy improvements with the advanced prompt, whereas Mixtral, GeminiPro, and GPT-3.5 Turbo see no such gains, suggesting that the additional details in the prompt may not benefit smaller or less sophisticated models.
Overall performance is quite poor with either prompting methods.

\noindent
\textbf{Performance gains with contextual information.} The inclusion of relevant examples via RAG significantly and consistently improves performance (Table~\ref{table:evaluation}, cf. RAG-top1/2/5 vs Prompt(advanced)). 
Notably, Mistral-m and GPT-4 Turbo exhibit marked improvements, suggesting they may possess a more advanced few-shot learning ability relative to the other models. 
Models outperform zero-shot prompting also when given a random dataset sample (RAG-random1), indicating that exposure to dataset structure and domain-specific language is helpful, even without query-specific context.

\noindent 
\textbf{Diminishing returns with increased context.} 
Providing a single example (RAG-top-1) leads to substantial improvements in performance, but adding more top results (RAG-top2 and RAG-top5) does not result in a similar increase. Some models exhibit a performance peak or a minor decline with additional context, indicating a limit to the beneficial amount of context.

\noindent
\textbf{Superiority of GPT-4 Turbo.} GPT-4 Turbo is the best model overall by a large margin, followed by GPT-3.5 Turbo. Mistral-m outperforms both Claude 2.1 and GeminiPro. The open-source Mixtral model lags behind proprietary models in both accuracy and executability across all scenarios.

\noindent
\textbf{Model-specific approach to oracle context.} 
In the RAG-top1-oracle scenario, where the prompt includes the correct SQL query, GPT-3.5 Turbo unexpectedly surpasses GPT-4 Turbo by closely mirroring the provided context, favouring direct replication. In contrast, GPT-4 Turbo and other models take a \quotes{deliberative} approach, often modifying the input, which, while useful for complex reasoning, hinders tasks that require exact copying.

\section{Related Work} \label{sec:Related}

\noindent 
\textbf{Text-to-SQL datasets for EHRs.} 
The MIMIC-SQL dataset ~\cite{wang-etal-2020-text} comprises 10\,000 template-generated questions for the MIMIC-III~\cite{johnson-etal-2016-mimic} database. 
It contains both question designed to retrieve patient-specific information, and questions on patients counts with logical and basic mathematical operations. 
\citet{tarbell2023-towards} noted limited diversity in MIMIC-SQL's queries, possibly affecting its utility for testing text-to-SQL model generalizability.
emrKBQA ~\cite{raghavan-etal-2021-emrkbqa} contains 1 million patient-specific questions, also based on MIMIC-III.
EHRSQL\cite{lee-2022-ehrsql} is a dataset created by extracting templates from clinical questions posed by hospital staff, which are then used to generate a comprehensive set of queries for MIMIC-III and eICU~\cite{pollard-etal-2018-eicu}. It relies on an earlier, less performing text-to-text model for query generation ~\cite{raffel-2020-exploring}.
All these datasets do not adhere with OMOP-CDM, and they opt for direct string matching for concept retrieval. The closest dataset to ours is the OMOP query library~\cite{ohdsi-2019-book,omop-query-library-2019}, which is a collection of queries in OMOP-CDM. 
We adapted and included fifteen SQL queries from this library pertinent to epidemiological research into our dataset.
\citet{park2023criteria2query} use rule-based methods and GPT-4 to translate clinical trial eligibility criteria into SQL queries for OMOP-CDM.
 
\noindent 
\textbf{Text-to-SQL with LLMs and in-domain demonstrations.} 
Prompting LLMs has proven effective, often outperforming specialized fine-tuned models in text-to-SQL task~\cite{pourreza-2023-dinsql}.
Both in-domain~\cite{chang2023prompt} and out-of-domain~\cite{chang-fosler-lussier-2023-selective} demonstrations improve LLMs' performance.
\citet{gao-2023-text} explores retrieval scenarios for in-domain demonstration selection. 
To the best of our knowledge, the exploration of these text-to-SQL methods within EHR (or biomedical) research has not yet extended to small datasets that are critical for industry applications.

\section{Conclusion} \label{sec:conclusion}
In this work, we presented the task of answering epidemiological questions using RWD.
We demonstrated that RAG is effective in improving performance on all tested scenarios. 
Our study extends the demonstrated efficacy of RAG from general text-to-SQL benchmarks~\cite{gao-2023-text, chang-fosler-lussier-2023-selective} to include to small, domain-specific biomedical datasets, underlining its utility in data-scarce industry settings.
The primary limitation is the dataset's limited size and specialized focus on epidemiological questions, suggesting further research should broaden its scope and scale.

\section{Acknowledgments} 
We would like to acknowledge Zuzanna Krawczyk-Borysiak for creating 27 text-SQL pairs; the remaining text-SQL pairs within the dataset were curated by A.Z.
We also thank Jared Worful, Melanie Hackl, and Zuzanna Krawczyk-Borysiak for helpful discussions.

\bibliography{anthology,custom}

\end{document}